\documentclass[letterpaper, 10 pt, conference]{ieeeconf}  

\IEEEoverridecommandlockouts                              

\usepackage{graphicx} 
\usepackage{epsfig} 
\usepackage{url}
\usepackage{hyperref}

\usepackage{fancyhdr}
\fancypagestyle{withfooter}{
  
  \fancyfoot[C]{\footnotesize Accepted to the IEEE ICRA Workshop on Field Robotics 2024}
}

\title{\LARGE \bf
Lessons Learned in Quadruped Deployment in Livestock Farming
}

\author{Francisco J. Rodríguez-Lera $^{*,1}$, Miguel   A. González-Santamarta $^{2}$,
 Jose Manuel Gonzalo Orden $^{2}$, \\
  Camino Fernández-Llamas $^{1}$, Vicente Matellán-Olivera
 and Lidia Sánchez-González $^{1}$ 
\thanks{$^{*}${Correspondence to: \tt\small fjrodl@unileon.es}}%
\thanks{$^{1}$Robotics Group, University of León, Spain}%
\thanks{$^{2}$Neurobiology Group, University of León, Spain}%
}

\begin{document}

\maketitle
\thispagestyle{withfooter}
\pagestyle{withfooter}

\begin{abstract}

The livestock industry faces several challenges, including labor-intensive management, the threat of predators and environmental sustainability concerns. Therefore, this paper explores the integration of quadruped robots in extensive livestock farming as a novel application of field robotics. The SELF-AIR project, an acronym for Supporting Extensive Livestock Farming with the use of Autonomous Intelligent Robots, exemplifies this innovative approach. Through advanced sensors, artificial intelligence, and autonomous navigation systems, these robots exhibit remarkable capabilities in navigating diverse terrains, monitoring large herds, and aiding in various farming tasks. This work provides insight into the SELF-AIR project, presenting the lessons learned.

\end{abstract}

\section{Introduction}

Extensive livestock farming plays a vital role in global food production, providing a significant portion of meat and dairy products consumed worldwide. However, this traditional agricultural practice is confronted with numerous challenges, including labor-intensive management, vast land areas to cover, and the need to ensure the well-being of the animals roaming freely across expansive pastures. Precision livestock farming involves the application of new technologies and intelligent sensors to help in daily tasks \cite{AQUILANI2022100429}: i.e. milking robots \cite{ARMSTRONG1997123}, virtual fencing \cite{MCSWEENEY2020105613}, ``Walk-over-Weigh'' infrastructures \cite{GONZALEZGARCIA2018226}, behavior monitoring for disease detection \cite{neethirajan2022automated}, among others. Herding is still a challenge due to the complexity of the tasks and it involves many aspects like context awareness, animal-robot-interaction, or autonomous navigation in the wild. For that reason, the existing solutions are proof of concepts analyzing the animal behavior \cite{ANZAI2022105751}, detecting cattle \cite{s18072048} or just theoretical approaches \cite{abbass2021shepherding,TSUNODA202310715}.

In response to these challenges, the integration of quadruped robots into extensive livestock farming has emerged as a promising solution to enhance productivity, animal welfare, and sustainability.

Inspired by the natural movement of animals, this project is called SELF-AIR which stands for Supporting Extensive Livestock Farming with the use of Autonomous Intelligent Robots, particularly the use of a quadruped robot. It offers unique capabilities suited for navigating diverse terrains and monitoring large herds in remote or rugged landscapes. By leveraging advanced sensors, artificial intelligence, and autonomous navigation systems, these robots can assist farmers in various tasks such as herd monitoring, health assessment, and environmental management. Moreover, their ability to operate autonomously for extended periods makes them well-suited for remote and challenging environments, where human intervention may be limited.

In this paper, we delve into the experience of deploying quadruped robots in extensive livestock farming, particularly exploring their potential to be them inside farming. Through a comprehensive examination of this case study, this paper aims to elucidate the opportunities and challenges associated with it.

The rest of the paper is presented as follows. Section\ref{sec:overall} presents the overall overview of the SELF-AIR project. Section\ref{sec:lessons} overviews our lessons learned. Finally, Section \ref{sec:conclusions} presents the conclusions of this work.

\section{Overall Overview}
\label{sec:overall}

When considering the deployment of a robotic platform on a sheep farm, some minimal skills and capabilities the robot should possess include:

\begin{enumerate}
\item Perception: The robot should be equipped with sensors such as cameras, LiDAR, or radar to perceive its surroundings. This includes detecting obstacles, terrain variations, and the presence of sheep and predators.

\item Navigation: The robot should be capable of navigating autonomously within the farm environment. This involves understanding its position relative to landmarks or GNSS (Global Navigation Satellite System) coordinates and planning efficient routes to reach designated areas. 


\item Path planning: The robot should be able to plan optimal paths for tasks such as herding sheep, monitoring pastures, or conducting inspections. This involves considering factors like terrain roughness, obstacles, and the location of the sheep.

\item Obstacle avoidance: The robot should possess the ability to detect and avoid obstacles in its path, such as rocks, fences, or other animals. This is crucial for ensuring the safety of both the robot and the livestock.

\item Interaction with sheep: Depending on the tasks assigned, the robot may need the capability to interact with sheep in a non-invasive manner. This could involve gentle herding techniques or monitoring the behavior and health of individual animals.

\item Communication: The robot should be able to communicate data and information to farm personnel or other systems in real-time. This may include transmitting images, sensor readings, or alerts about detected anomalies.

\item Endurance: Given the extensive nature of sheep farms, the robot should have sufficient battery life or energy autonomy to operate for extended periods without frequent recharging or refueling.

\item Adaptability to environmental conditions: The robot should be able to operate effectively in various environmental conditions commonly encountered on a sheep farm, including uneven terrain, inclement weather, and low light conditions.

\end{enumerate}

\begin{figure}[ht]
\centering
\includegraphics[width=0.5\textwidth]{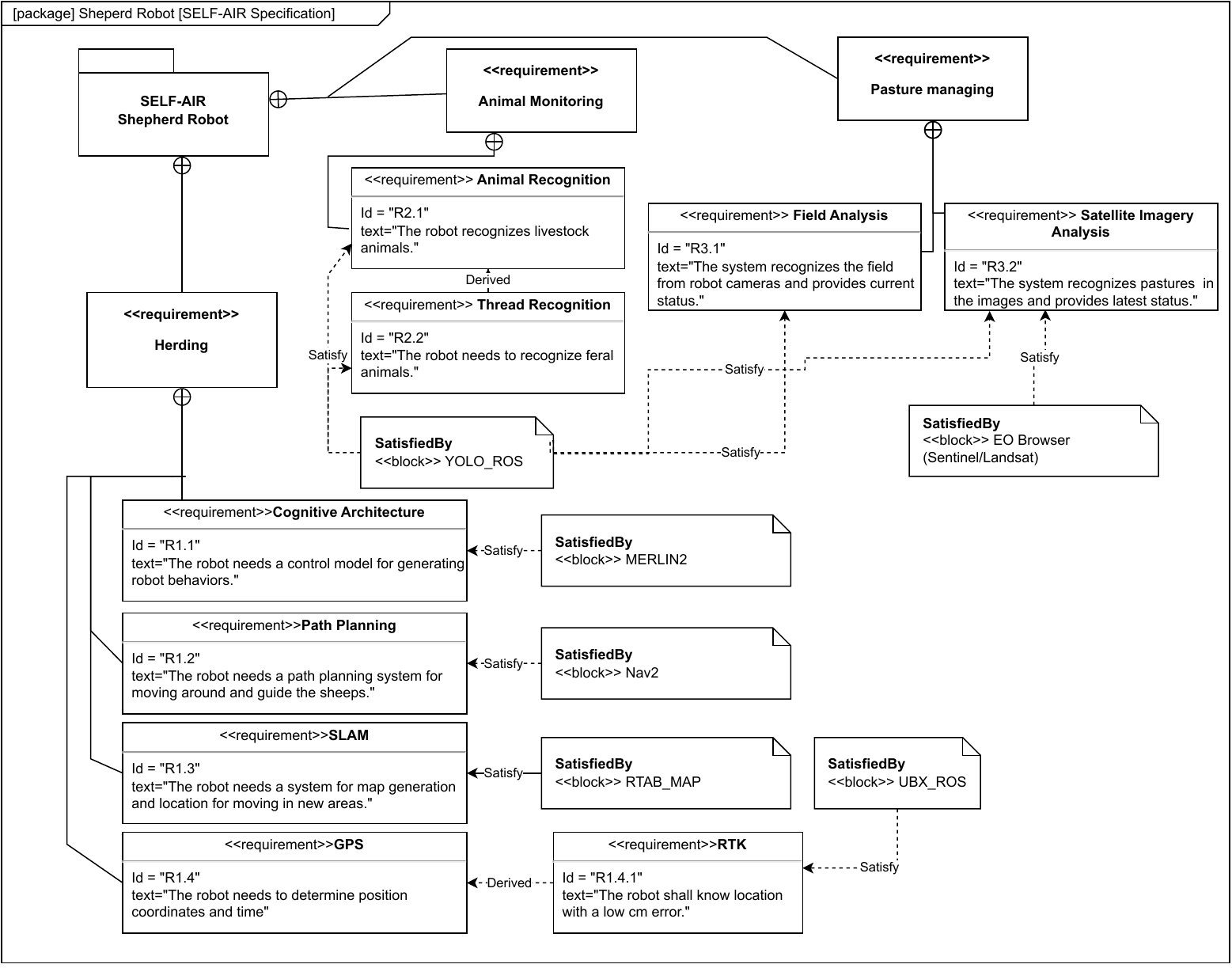}
\caption{SELF-AIR project requirements (first iteration) .\label{fig:selfAirRequeriment} } 
\end{figure}

These components can be joined in a set of requirements needed for fulfilling the main tasks described in the project proposal:
1) herding, a mechanism for moving the flock around, 2) predator monitoring, and 3) pasture managing. Fig~\ref{fig:selfAirRequeriment} presents the set of requirements for reaching some of the goals. 

\begin{itemize}
    \item[R1.1] Cognitive Architecture: The robot needs a control model for generating robot behaviors. In this research, we use MERLIN2~\cite{GONZALEZSANTMARTA2023100477}, a hybrid cognitive architecture integrated into ROS 2 \cite{macenski2022robot} and designed for autonomous robots. MERLIN2 uses symbolic knowledge to model the robot's world, based on PDDL \cite{PDDL}; a deliberative system for planning to achieve goals, that uses PDDL planners such as POPF \cite{popf}; state machines for immediate behaviors, created with YASMIN \cite{yasmin}; and emergent modules for object recognition, and both speech recognition and synthesis.
    
    \item[R1.2] Path Planning: The robot needs a path planning system for moving around and guiding the sheep. Thus, navigation can be handled by using solutions like Navigation2 \cite{macenski2020marathon2}, which is the navigation stack for ROS 2.
    
    \item[R1.3] SLAM The robot needs a system for map generation and location for moving to new areas. One way for a robot to achieve localization is through odometry~\cite{borenstein1996sensors} which can be improved using cameras' images to perform visual odometry~\cite{nister2004visual}. Moreover, Kalman filters \cite{moore2016generalized} can be used to fusion the odometry with other sensor data, such as IMU data. Finally, Simultaneous Localization and Mapping (SLAM) \cite{durrant2006simultaneous,thrun2007simultaneous} can be applied to produce maps used by the robots to navigate and localize. As a result, a map is created while the robot is simultaneously attempting to localize itself using that obtained. Visual SLAM (VSLAM)~\cite{fuentes2015visual}, which can performed using tools like RTAB-Map \cite{labbe2019rtab}, is useful in the case presented in this due to creating maps with the cameras' images.
    
    \item[R1.4] GPS: The robot needs to determine the global position coordinates and time using GNSS (Global Navigation Satellite System) satellites.
    
    \item[R1.4.1] GPS+RTK: The robot shall know the location with a low cm error by using GNSS data along with RTK (Real-Time Kinematics) \cite{gan2007implement} to obtain accurate global positioning.
    
    \item[R2.1] Animal recognition: The robot recognizes livestock animals.  To treat this, object detection tools can be employed, for instance, YOLOv8 \cite{Jocher_YOLO_by_Ultralytics_2023,yolov8_ros_2023} can be trained and used to detect sheep.
    
    \item[R2.2] Threat recognition: The robot needs to recognize feral animals. Previous work shows promising results using YOLO models to provide the capability to distinguish between prey and potential predators to 4-legged robots \cite{mdpiLobos2022}.
    
    \item[R3.1] Field analysis: The system recognizes the field from robot cameras and provides the current status.
    
    \item[R3.2] Satellite Imagery analysis: The system recognizes pastures in the images and provides the latest status from the satellite database. 
\end{itemize}

The requirements presented are just the initial revision in order to select and prepare the robot for the field. 

\subsection{Hardware}

Quadruped robots mimic the locomotion of animals with four legs, enabling them to navigate diverse terrains, including rugged or uneven surfaces commonly found in agricultural environments. This project employs two different platforms, the Unitree A1 platform \cite{unitree} and Ghost Vision 60~\cite{ghostroboticsVISIONGhost}. It highlights the main features of the robot based on physical shape, autonomy, maneuverability, 

\subsubsection{Unitree A1}

The size of this robot is about 500 mm (length) 300 mm (wide) 400 mm (tall) while its weight is 12 kg. Besides, it is capable of running at a maximum speed of 3.3 m/s. Moreover, the maximum torque is 33.5 Nm. Finally, its runtime is between 1 and 2.5 hours.

\subsubsection{Ghost Vision 60}

The size of this robot is about 850 mm (length) 540 mm (wide) 760 mm (tall) while its weight is 51 kg. Besides, it is capable of running at a maximum speed of 3.0 m/s. Moreover, the maximum torque is 105.0 Nm. Finally, its runtime is between 8 and 10 hours.

\begin{figure}[ht]
\centering
\includegraphics[width=0.5\textwidth]{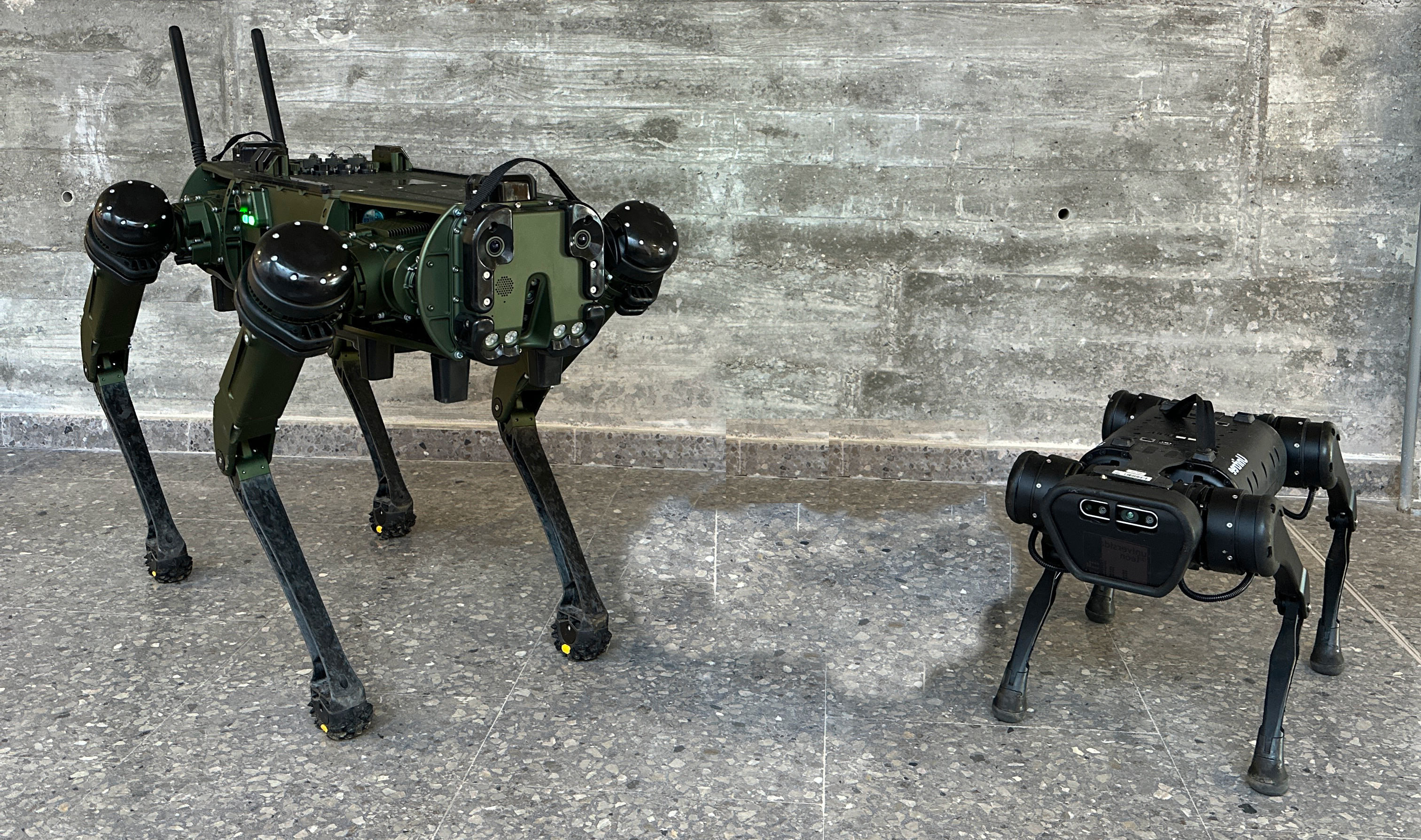}
\caption{Ghost Vision-60 (left) and Unitree A1 (right).\label{fig:robots} } 
\end{figure}

\subsection{Field Tests}

The initial tests, recording and data presented here was developed with a remote operator. 

\subsubsection{Small-size Farm - Private - Quadruped Robot A1 - 2 Sheep}

In this experiment conducted on a private farm, we aim to explore the dynamics of herding using a small robot (Unitree A1) and two sheep. The experiment will be set in a controlled environment to closely observe how the robot interacts with the sheep and influences their movement patterns. By utilizing the robot as a herding agent, we seek to understand its effectiveness in guiding the sheep toward a desired location within the farm. Various parameters such as the robot's speed, proximity to the sheep, and behavioral responses of the sheep will be monitored and analyzed. 
There is a video of the experiment presented here~\url{https://youtu.be/jCZgwwRknMk?si=9RN-EfI2YN5mnO1L}.


\subsubsection{Small-size Research Farm - UNEX - Quadruped Robot A1 + Drone - 22 Sheep}

In this innovative experiment conducted at Extremadura  University's research farm, we are investigating the dynamics of herding using a combination of a small-sized robot, 22 sheep, and a drone. The primary objective of this study is to track the flock's movements and behavior patterns in real-time, leveraging the capabilities of both robotics and aerial surveillance technology. By deploying the small-sized robot alongside the drone, we aim to observe how the quadruped agent guides. This interdisciplinary robotic approach not only explores the integration of robotics and drones in livestock management but also holds promising implications for enhancing the efficiency and welfare standards of agricultural practices, locally, without satellite imagery. 
There is a video of the experiment presented here~\url{https://youtu.be/bdnpUM-3siU?si=OaClAui9IdmdlnaX}.


\subsubsection{Medium-size Farm - Private - No Robot - 300 Sheep}

In this experimental study set on a medium-sized private farm, we delve into the intricate dynamics of herding without the use of robots, instead employing a goat on a farm of around 300 sheep. The experiment aims to explore the natural herding instincts and behaviors exhibited by these animals when placed in a group setting. By observing their interactions and movements within the farm's confines, we aim to gain insights into the emergent patterns of collective behavior, leadership dynamics, and the roles played by individual animals, particularly the goats, known for their independent nature. This study not only contributes to our understanding of herd behavior in mixed-species groups but also provides valuable insights for optimizing herding strategies in agricultural contexts where the use of robots may not be feasible or desirable.
There is a video of the experiment presented here~\url{https://youtu.be/EhTT-AQ6GpU}.




\subsubsection{Medium-size Research Farm - ULE - single Robot - 50 Sheep}

In this experimental study conducted within the confines of León University's research farm, we investigated the dynamics of herding using a significant robotic presence (Vision 60) alongside 100 sheep. By deploying a large-sized robot as the primary herding agent, we aim to explore its efficacy in guiding and managing the movement of the sheep within the experimental farm. This research seeks to uncover the potential advantages and challenges associated with robotic herding, such as the robot's ability to navigate diverse terrain, maintain proximity to the sheep, and influence their collective behavior. Additionally, we aim to assess the welfare implications for the animals involved and the overall efficiency of utilizing robotic technology in agricultural practices. Through meticulous observation and analysis, this experiment promises to offer valuable insights into the future integration of robotics within livestock management systems, contributing to advancements in both animal welfare and agricultural efficiency. There is a video of the experiment presented here~\url{https://youtu.be/onak2nhyCx8}.



\begin{figure*}[ht]
\centering
\includegraphics[width=0.23\textwidth]{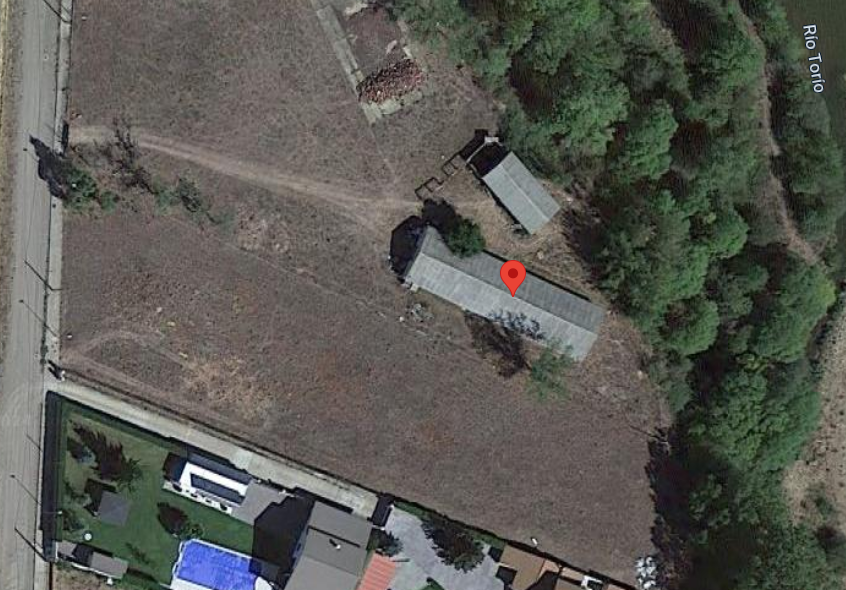}
\includegraphics[width=0.2\textwidth]{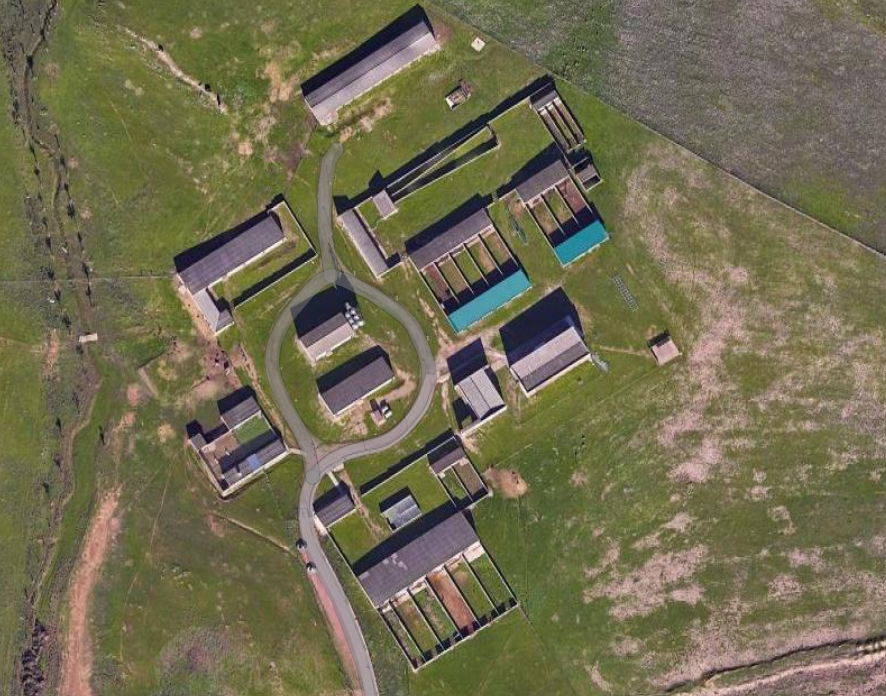}
\includegraphics[width=0.24\textwidth]{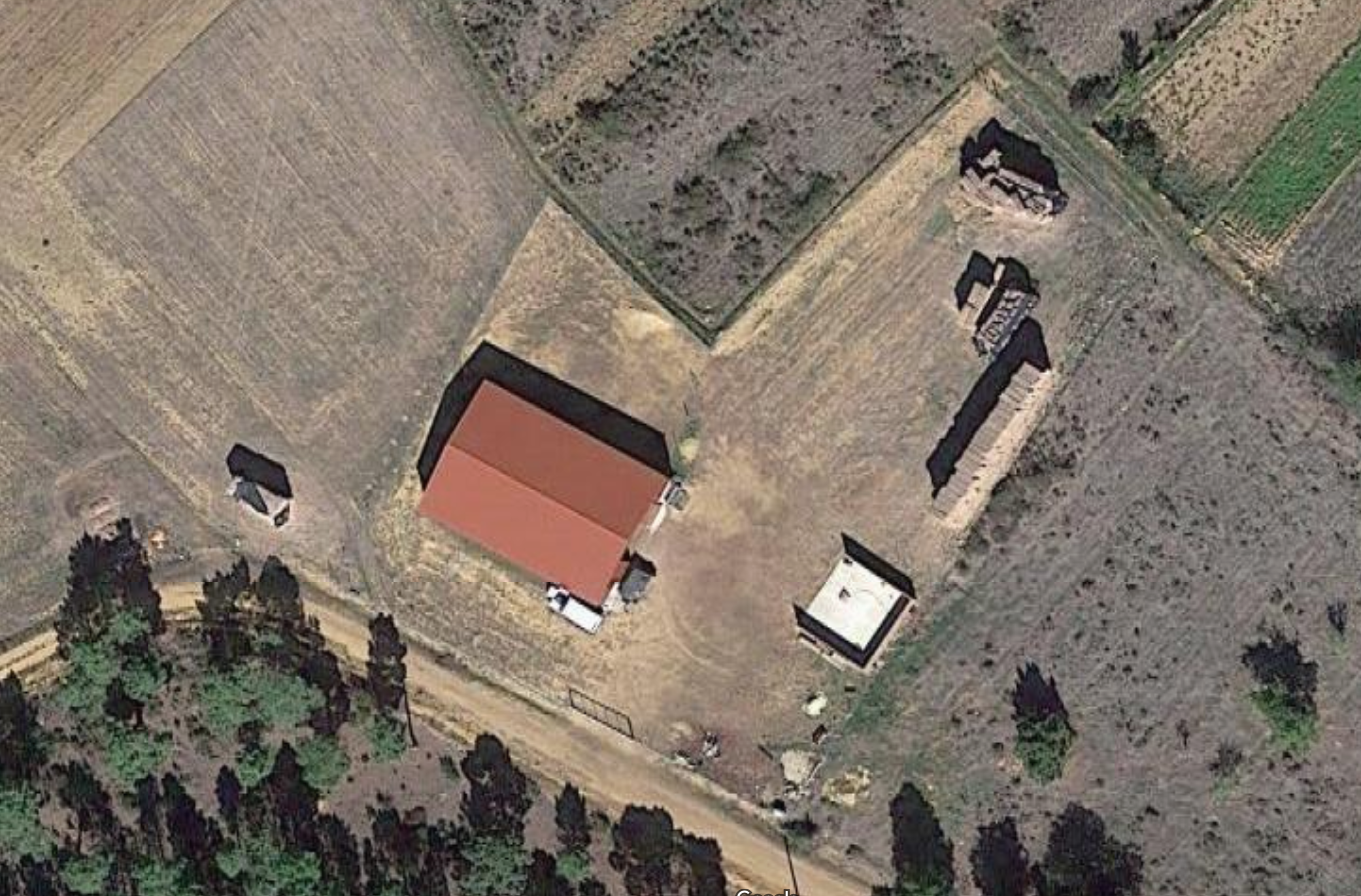}
\includegraphics[width=0.24\textwidth]{figures/GranjaExperimental\_Leon.png}
\caption{Google Maps view of four farms visited on this research.\label{fig:granjas} } 
\end{figure*}

\section{Lessons Learned}
\label{sec:lessons}

Through the process of deploying a robotic platform on three different sheep farms, we have gained valuable insights and lessons learned. 
The next items will introduce the advantages and disadvantages found:

- Advantages
   \begin{itemize}
        \item  \textbf{Livestock Monitoring:} Quadrupeds equipped with sensors can enhance precision in monitoring the health and behavior of sheep, facilitating early detection of diseases or anomalies.
        \item  \textbf{Pasture Navigation:} Quadrupeds are well-suited for navigating through uneven pastures, ensuring effective monitoring and management of sheep across varying terrains.
        \item  \textbf{Extended Patrols:} Energy-efficient quadrupeds can perform longer patrols, enhancing their ability to monitor large grazing areas.
        \item  \textbf{Efficient Herding:} Quadrupeds can assist in herding and managing the flock, reducing the manual effort required by human shepherds.
         
         \item \textbf{Behavioral Insights:} Integration of data analytics allows for in-depth analysis of sheep behavior, aiding in informed decision-making for health management and breeding programs.

        \item \textbf{All-Weather Monitoring:} Quadrupeds designed for weather resistance can monitor sheep in various weather conditions, ensuring continuous observation.

         \item \textbf{Remote Herding Control:} Remote monitoring enables shepherds to control quadrupeds, assisting in herding and managing the flock from a distance.

         \item   \textbf{Enhanced Shepherd Skills:} Training shepherds to work alongside quadrupeds enhances their skills in managing and utilizing the technology effectively.

        \item    \textbf{Legal Compliance:} Adhering to regulations ensures the ethical and lawful use of quadrupeds in sheep farming.

        \item \textbf{Increased Efficiency in Herding:} A thorough cost-benefit analysis can reveal increased efficiency in herding and monitoring, leading to potential long-term economic benefits.

          \item \textbf{Flexible Integration:} Quadrupeds can be designed to integrate seamlessly with existing sheep farming practices, offering a scalable solution.

        \item   \textbf{Community Engagement:} Engaging with the local sheep farming community and addressing concerns promotes acceptance and support for the technology.
          
      \end{itemize}

- Disadvantages
        
   \begin{itemize}
        \item \textbf{Animal Interaction:} Quadrupeds might need a careful introduction to the flock to avoid causing stress or disturbance among the sheep.
        
        \item  \textbf{Flock Disturbance:} Care must be taken to minimize any disturbance caused by the quadrupeds, as they move through the pasture with the flock.
        
        \item  \textbf{Limited Endurance:} Energy constraints may limit the duration of patrols, requiring strategic deployment and recharging schedules.
        
        \item \textbf{Herding Instinct:} Quadrupeds may need careful training to exhibit effective herding behaviors without causing stress to the sheep.

        \item  \textbf{Data Privacy:} Ensuring the privacy and security of sensitive data related to individual sheep is crucial to prevent unauthorized access.

         \item \textbf{Training Requirements:} Effective collaboration requires training both the quadrupeds and shepherds to work together seamlessly.

         \item  \textbf{Equipment Protection:} Adequate protection of electronic components is necessary to prevent damage from exposure to rain or extreme temperatures.

         \item \textbf{Signal Range:} The range of remote control signals may be limited, requiring careful consideration of the operational distance.

         \item \textbf{Resistance to Change:} Resistance to adopting new technologies among shepherds may pose challenges, necessitating comprehensive training programs.

        \item \textbf{Regulatory Restrictions:} Stringent regulations may limit certain activities or capabilities of quadrupeds in livestock management.

        \item   \textbf{Initial Investment Burden:} High upfront costs may pose a financial burden, requiring careful consideration of the long-term returns.

        \item    \textbf{Operational Adjustment:} Some adjustments in farm practices may be needed to optimize the integration of quadrupeds into the existing workflow.

        \item   \textbf{Cultural Resistance:} Traditional farming communities may be resistant to change, necessitating efforts to showcase the benefits and address cultural concerns.
        
   \end{itemize}

\section{Conclusions}
\label{sec:conclusions}

In conclusion, our study sheds light on the significant potential of quadruped robots in livestock farming practices. Through our experimentation and observations, we have gleaned valuable insights into the benefits and challenges associated with deploying these robots in agricultural settings, particularly with livestock farms. Our findings underscore the importance of careful planning, robust design, and continuous adaptation to effectively integrate quadruped robots into livestock management routines. Despite encountering hurdles such as terrain navigation and animal acceptance, our study demonstrates the promising role of robotics in enhancing efficiency, productivity, and animal welfare on farms. Moving forward, further research and development efforts are warranted to refine the autonomous capabilities of quadruped robots and optimize their performance in real-world farming environments.




\section*{ACKNOWLEDGMENT}
Grant TED2021-132356B-I00 funded by MCIN/AEI/10.13039/501100011033 and by the ``European Union NextGenerationEU/PRTR''.
We thank the drone operators, quadruped robot operators, laboratory technicians, directors, and farm managers, as well as private shepherds, for their assistance during these initial phases of the research.



\bibliographystyle{unsrt}
\bibliography{root}

\begin{thebibliography}{10}

\bibitem{AQUILANI2022100429}
C.~Aquilani, A.~Confessore, R.~Bozzi, F.~Sirtori, and C.~Pugliese.
\newblock Review: Precision livestock farming technologies in pasture-based
  livestock systems.
\newblock {\em Animal}, 16(1):100429, 2022.

\bibitem{ARMSTRONG1997123}
D.V. Armstrong and L.S. Daugherty.
\newblock Milking robots in large dairy farms.
\newblock {\em Computers and Electronics in Agriculture}, 17(1):123--128, 1997.
\newblock Robotic Milking.

\bibitem{MCSWEENEY2020105613}
Diarmuid McSweeney, Bernadette O'Brien, Neil~E. Coughlan, Alexis Férard,
  Stepan Ivanov, Paddy Halton, and Christina Umstatter.
\newblock Virtual fencing without visual cues: Design, difficulties of
  implementation, and associated dairy cow behaviour.
\newblock {\em Computers and Electronics in Agriculture}, 176:105613, 2020.

\bibitem{GONZALEZGARCIA2018226}
E.~González-García, M.~Alhamada, J.~Pradel, S.~Douls, S.~Parisot,
  F.~Bocquier, J.B. Menassol, I.~Llach, and L.A. González.
\newblock A mobile and automated walk-over-weighing system for a close and
  remote monitoring of liveweight in sheep.
\newblock {\em Computers and Electronics in Agriculture}, 153:226--238, 2018.

\bibitem{neethirajan2022automated}
Suresh Neethirajan.
\newblock Automated tracking systems for the assessment of farmed poultry.
\newblock {\em Animals}, 12(3):232, 2022.

\bibitem{ANZAI2022105751}
Hiroki Anzai and Hina Sakurai.
\newblock Preliminary study on the application of robotic herding to
  manipulation of grazing distribution: Behavioral response of cattle to
  herding by an unmanned vehicle and its manipulation performance.
\newblock {\em Applied Animal Behaviour Science}, 256:105751, 2022.

\bibitem{s18072048}
Alberto Rivas, Pablo Chamoso, Alfonso González-Briones, and Juan~Manuel
  Corchado.
\newblock Detection of cattle using drones and convolutional neural networks.
\newblock {\em Sensors}, 18(7), 2018.

\bibitem{abbass2021shepherding}
Abbass~Hussein A. and Robert~A. Hunjet, editors.
\newblock {\em Shepherding UxVs for Human-Swarm Teaming. An Artificial
  Intelligence Approach to Unmanned X Vehicles}.
\newblock Springer Cham, 2021.
\newblock \url{https://doi.org/10.1007/978-3-030-60898-9}.

\bibitem{TSUNODA202310715}
Yusuke Tsunoda, Teruyo Wada, and Koichi Osuka.
\newblock Proposal of general shepherding controller for global stability:
  Backstepping technique approach.
\newblock {\em IFAC-PapersOnLine}, 56(2):10715--10720, 2023.
\newblock 22nd IFAC World Congress.

\bibitem{GONZALEZSANTMARTA2023100477}
Miguel~\'{A}. Gonz\'{a}lez-Santamarta, Francisco~J. Rodr\'{i}guez-Lera, Camino
  Fernández-Llamas, and Vicente Matell\'{a}n-Olivera.
\newblock Merlin2: Machined ros 2 planing.
\newblock {\em Software Impacts}, 15:100477, 2023.

\bibitem{macenski2022robot}
Steven Macenski, Tully Foote, Brian Gerkey, Chris Lalancette, and William
  Woodall.
\newblock Robot operating system 2: Design, architecture, and uses in the wild.
\newblock {\em Science Robotics}, 7(66):eabm6074, 2022.

\bibitem{PDDL}
Maria Fox and Derek Long.
\newblock Pddl2.1: An extension to pddl for expressing temporal planning
  domains.
\newblock {\em J. Artif. Intell. Res. (JAIR)}, 20:61--124, 12 2003.

\bibitem{popf}
Amanda Coles, Andrew Coles, Maria Fox, and Derek Long.
\newblock Forward-chaining partial-order planning.
\newblock In {\em ICAPS 2010 - Proceedings of the 20th International Conference
  on Automated Planning and Scheduling}, pages 42--49, 01 2010.

\bibitem{yasmin}
Miguel~{\'A}. Gonz{\'a}lez-Santamarta, Francisco~J. Rodr{\'i}guez-Lera, Vicente
  Matell{\'a}n-Olivera, and Camino Fern{\'a}ndez-Llamas.
\newblock {YASMIN}: Yet another state machine.
\newblock In Danilo Tardioli, Vicente Matell{\'a}n, Guillermo Heredia,
  Manuel~F. Silva, and Lino Marques, editors, {\em ROBOT2022: Fifth Iberian
  Robotics Conference}, pages 528--539, Cham, 2023. Springer International
  Publishing.

\bibitem{macenski2020marathon2}
Steve Macenski, Francisco Martín, Ruffin White, and Jonatan Ginés~Clavero.
\newblock The marathon 2: A navigation system.
\newblock In {\em 2020 IEEE/RSJ International Conference on Intelligent Robots
  and Systems (IROS)}, 2020.

\bibitem{borenstein1996sensors}
Johann Borenstein, HR~Everett, Liqiang Feng, et~al.
\newblock Where am i? sensors and methods for mobile robot positioning.
\newblock {\em University of Michigan}, 119(120):27, 1996.

\bibitem{nister2004visual}
David Nist{\'e}r, Oleg Naroditsky, and James Bergen.
\newblock Visual odometry.
\newblock In {\em Proceedings of the 2004 IEEE Computer Society Conference on
  Computer Vision and Pattern Recognition, 2004. CVPR 2004.}, volume~1, pages
  I--I. Ieee, 2004.

\bibitem{moore2016generalized}
Thomas Moore and Daniel Stouch.
\newblock A generalized extended kalman filter implementation for the robot
  operating system.
\newblock In {\em Intelligent autonomous systems 13}, pages 335--348. Springer,
  2016.

\bibitem{durrant2006simultaneous}
Hugh Durrant-Whyte and Tim Bailey.
\newblock Simultaneous localization and mapping: part i.
\newblock {\em IEEE robotics \& automation magazine}, 13(2):99--110, 2006.

\bibitem{thrun2007simultaneous}
Sebastian Thrun.
\newblock Simultaneous localization and mapping.
\newblock In {\em Robotics and cognitive approaches to spatial mapping}, pages
  13--41. Springer, 2007.

\bibitem{fuentes2015visual}
Jorge Fuentes-Pacheco, Jos{\'e} Ruiz-Ascencio, and Juan~Manuel
  Rend{\'o}n-Mancha.
\newblock Visual simultaneous localization and mapping: a survey.
\newblock {\em Artificial intelligence review}, 43(1):55--81, 2015.

\bibitem{labbe2019rtab}
Mathieu Labb{\'e} and Fran{\c{c}}ois Michaud.
\newblock Rtab-map as an open-source lidar and visual simultaneous localization
  and mapping library for large-scale and long-term online operation.
\newblock {\em Journal of Field Robotics}, 36(2):416--446, 2019.

\bibitem{gan2007implement}
Samuel Gan-Mor, Rex~L Clark, and Bruce~L Upchurch.
\newblock Implement lateral position accuracy under rtk-gps tractor guidance.
\newblock {\em Computers and Electronics in Agriculture}, 59(1-2):31--38, 2007.

\bibitem{Jocher_YOLO_by_Ultralytics_2023}
Glenn Jocher, Ayush Chaurasia, and Jing Qiu.
\newblock {YOLO by Ultralytics}, January 2023.

\bibitem{yolov8_ros_2023}
Miguel~Á. González-Santamarta.
\newblock {yolov8\_ros}, February 2023.

\bibitem{mdpiLobos2022}
Virginia Riego~del Castillo, Lidia Sánchez-González, Adrián Campazas-Vega,
  and Nicola Strisciuglio.
\newblock Vision-based module for herding with a sheepdog robot.
\newblock {\em Sensors}, 22(14), 2022.

\bibitem{unitree}
Unitree a1.
\newblock \url{https://m.unitree.com/a1/}.
\newblock [Accessed 19-03-2024].

\bibitem{ghostroboticsVISIONGhost}
{V}{I}{S}{I}{O}{N} 60 | {G}host {R}obotics --- ghostrobotics.io.
\newblock \url{https://www.ghostrobotics.io/vision-60}.
\newblock [Accessed 19-03-2024].

\end{thebibliography}

\addtolength{\textheight}{-11cm}   

\end{document}